\documentclass[runningheads]{llncs}
\usepackage[T1]{fontenc}
\usepackage{graphicx,verbatim}
\usepackage[T1]{fontenc}
\usepackage{graphicx}
\usepackage{amsmath}
\usepackage{amssymb}
\usepackage{booktabs}
\usepackage{multirow}
\usepackage{xcolor}
\usepackage{microtype}

\begin{document}
\title{Compositional Cross-Modality Translation via Whole-Volume Multitask Latent Flow Matching}
\titlerunning{Whole-Volume Multitask Latent Flow Matching}

\author{Daniele Molino\inst{1}\thanks{Corresponding author.}
\and
Alessio Zoboli\inst{1}
\and
Camillo Maria Caruso\inst{1}
\and
Valerio Guarrasi\inst{3}
\and
Paolo Soda\inst{1,2}
}

\authorrunning{D. Molino et al.}
\institute{Unit of Artificial Intelligence and Computer Systems, Department of Engineering, Università Campus Bio-Medico di Roma, Rome, Italy\\
\email{daniele.molino@unicampus.it, alessio.zoboli@alcampus.it, camillomaria.caruso@unicampus.it, p.soda@unicampus.it}
\and
Department of Diagnostics and Intervention, Biomedical Engineering and Radiation Physics, Umeå University, Umeå, Sweden\\
\email{paolo.soda@umu.se}
\and
UniCamillus -- Saint Camillus International University of Health Sciences, Rome, Italy\\
\email{valerio.guarrasi@unicamillus.org}}
  
\maketitle              

\begin{abstract}
Cross-modality medical image translation can reduce the burden of multi-modal acquisitions, yet the field remains constrained by two coupled limitations: methods operate on 2D slices or 3D patches rather than whole volumes, and train a separate model for each translation task. 
Both stem from a single cause, the absence of a sufficiently strong volumetric prior, which forces generative models to learn anatomical appearance and cross-modality mapping simultaneously, an ill-posed problem at the scale of available paired datasets. 
We propose to decouple these objectives. 
A large-scale pretrained 3D variational autoencoder provides a compact latent representation of volumetric appearance, reducing translation to a conditional flow-matching problem.
This compression makes whole-volume processing tractable, while a resolution-aware sampling strategy preserves native anatomical scale. 
We train a single model jointly across inter-modality (MRI$\to$CT, CBCT$\to$CT) and intra-modality (MRI$\to$MRI) tasks over three multi-center datasets. 
Across all tasks, whole-volume processing outperforms its patch-based counterpart, and the multi-task model matches task-specific baselines while replacing $N$ networks with one. 
Crucially, joint training unlocks capabilities inaccessible to task-specific approaches: zero-shot generalization to anatomical regions unseen during training, within 0.15 SSIM of the fully supervised model, and compositional cross-dataset translation along paths never directly supervised.
These results suggest that combining a strong volumetric prior with multitask training is a scalable route toward synthesis systems that generalize beyond their training distribution.
Code is available at \url{https://github.com/arco-group/Whole-Volume-Latent-FM}.

\end{abstract}

\section{Introduction}
Cross-modality medical image translation has emerged as a clinically relevant paradigm for reducing the burden of multi-modal acquisitions.
By synthesizing a target modality directly from an available source, it enables applications ranging from MR-only radiotherapy planning~\cite{thummerer2023} to scanner harmonization and data augmentation under strict privacy constraints~\cite{dayarathna2024,chen2025,di2025texture}.
The growing clinical interest in this problem has catalyzed dedicated challenges and benchmarks.
SynthRad2023 and SynthRad2025~\cite{thummerer2023,thummerer2025} provide multi-center paired datasets spanning five anatomical regions, covering both magnetic resonance imaging (MRI)--computed tomography (CT) and cone-beam CT (CBCT)--CT translation tasks. BraTS2023~\cite{baid2021} complements these with multi-parametric MRI acquisitions for brain tumour cases, enabling intra-modality translation across four MRI sequences.
Yet paired datasets remain scarce relative to the scale of modern deep learning: matched multi-modal acquisitions require coordinated imaging protocols and expose patients to additional procedures, while data-sharing regulations strictly limit cross-institutional pooling~\cite{kaissis2020secure,kazerouni2023}.
Despite progress from GAN-based approaches~\cite{goodfellow2014,isola2017,zhu2017} to diffusion and flow matching models~\cite{ho2020,lipman2022,rombach2022}, most medical image translation methods still operate on 2D slices or 3D patches~\cite{romoli2026}.
This is largely driven by the cubic memory and compute cost of full-volume generation~\cite{niyas2022medical}. 
However, patch-based models lack global anatomical context, may introduce stitching artifacts~\cite{molino2025text,molino2026retrieval}, and often require resampling to a common voxel grid, thereby altering native scale relationships. 
Additionally, most approaches train one model per translation task, limiting representation sharing across modalities, anatomical regions, and datasets. 
Since each model is trained on a single narrow distribution, it cannot exploit the regularities shared across modalities, anatomical regions, and acquisition centers~\cite{iele2025tta,glocker2019scanner}.
Furthermore, the task-specific formulation structurally precludes generalization to modality compositions absent from any individual training set, a capability that only emerges from joint training over shared representations~\cite{molino2026xgem,molino2025any}.
We argue that these limitations stem not from the translation task itself but from the absence of a strong volumetric prior: when a model must simultaneously learn how medical volumes look \emph{and} how to translate between them, the problem is ill-posed at the scale of available paired datasets. 
We address these limitations by decoupling volumetric representation learning from cross-modality mapping. 
A pretrained 3D VAE~\cite{guo2025maisi}, trained on over 55,000 CT and MRI volumes, maps whole scans to a compact latent space.
Translation is then learned as a conditional flow-matching problem in this latent space, making whole-volume processing tractable.
On top of this representation, we train a conditional flow-matching model~\cite{lipman2022} jointly across all available modalities, anatomical regions, and datasets, using a resolution-aware bucketed sampling strategy that preserves each volume's native resolution without resampling.
Our contributions are threefold. 
\textbf{(i)} We show that whole-volume latent flow matching outperforms its patch-based counterpart, empirically validating that global context improves translation quality. 
\textbf{(ii)} We train a single multitask model across MRI$\to$CT, CBCT$\to$CT, and MRI$\to$MRI translation, matching task-specific baselines without significant degradation. 
\textbf{(iii)} We demonstrate that joint training across datasets unlocks capabilities structurally inaccessible to task-specific approaches, such as zero-shot generalization to anatomical regions unseen during training, and compositional translation across modalities and datasets never co-observed during training.

\begin{figure}[t]
  \centering
  \includegraphics[width=\textwidth]{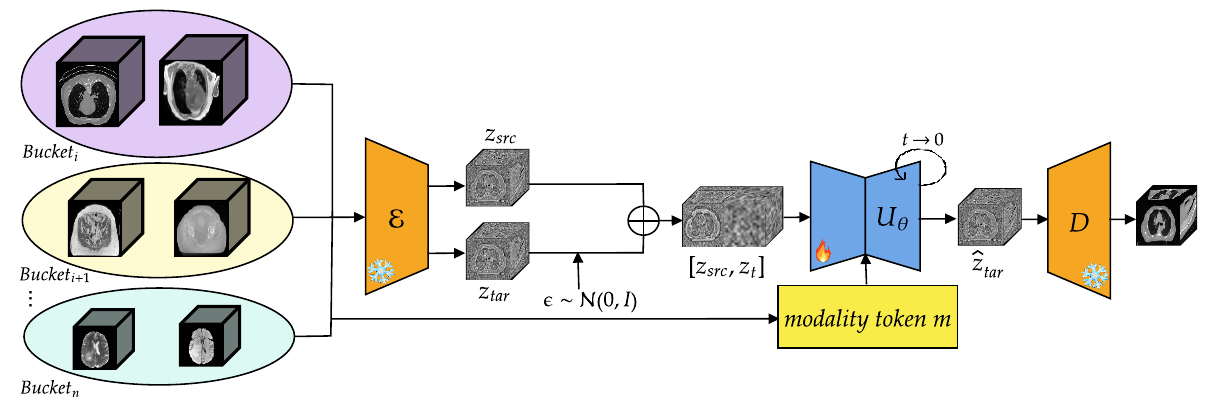}
  \caption{Overview of the proposed multitask whole-volume framework. Native-resolution volumes are processed through bucketed sampling and encoded by a frozen pretrained 3D VAE into a compact latent space. A target-modality token conditions the latent flow-matching model, enabling both direct cross-modality translation and compositional inference through chained translation steps.}
  \label{fig:framework}
\end{figure}

\section{Method}
The proposed framework comprises two stages, latent compression and latent translation, reflecting a deliberate decoupling of volumetric representation learning from cross-modality mapping.
An overview is provided in Fig.~\ref{fig:framework}.

\paragraph{Volumetric latent compression.}
We adopt the frozen encoder $\mathcal{E}(\cdot)$ of MAISI~\cite{guo2025maisi}, a 3D VAE pretrained on over 55{,}000 heterogeneous CT and MRI volumes.
Given an input volume $\mathbf{x} \in \mathbb{R}^{H \times W \times D}$, the
encoder produces a compact latent representation via the reparameterization trick:
\begin{equation}
  \mathbf{z} = \boldsymbol{\mu}_\phi(\mathbf{x})
  + \boldsymbol{\sigma}_\phi(\mathbf{x}) \odot \boldsymbol{\epsilon},
  \quad \boldsymbol{\epsilon} \sim \mathcal{N}(\mathbf{0},\mathbf{I}),
  \quad \mathbf{z} \in \mathbb{R}^{h \times w \times d \times c},
\end{equation}
with $h \ll H$, $w \ll W$, $d \ll D$.
The decoder $\mathcal{D}(\cdot)$ defines the inverse mapping $\tilde{\mathbf{x}} = \mathcal{D}(\mathbf{z})$. 
Both are kept frozen, confining translation to a latent space that already encodes the complexity of medical volumes. 

\paragraph{Conditional latent translation via flow matching.}
Given source and target latents $\mathbf{z}^{\text{src}} = \mathcal{E}(\mathbf{x}^{\text{src}}), \mathbf{z}^{\text{tar}}=\mathcal{E}(\mathbf{x}^{\text{tar}})$, we define a probability path via linear interpolation:
\begin{equation}
  \mathbf{z}_t = (1-t)\,\mathbf{z}^{\text{tar}}
  + t\,\boldsymbol{\epsilon},
  \quad \boldsymbol{\epsilon} \sim \mathcal{N}(\mathbf{0},\mathbf{I}),
  \quad t \in [0,1],
\end{equation}
with target velocity field $u_t = \boldsymbol{\epsilon} - \mathbf{z}^{\text{tar}}$. 
A 3D U-Net~\cite{ronneberger2015} $\mathcal{U}_\theta(\cdot)$ is trained to approximate this field conditionally on $\mathbf{z}^{\text{src}}$ and a target modality token $m \in \mathcal{M}$, where $\mathcal{M}$ is the set of modalities spanned by the training data; the source latent is injected via channel-wise concatenation, while $m$ is projected through a learnable embedding and combined with the timestep encoding.
No source-modality token is used, as the source modality is already specified by $\mathbf{z}^{\mathrm{src}}$, while the target token disambiguates the modality to be synthesized.
The model is optimized by minimizing the flow matching objective:
\begin{equation}
  \mathcal{L}_{\text{FM}}
  = \mathbb{E}_{(z^{src},z^{tar}),\epsilon,t}
  \bigl[\bigl\|
  v_\theta(\mathbf{z}_t, t, \mathbf{z}^{\text{src}}, m)
  - (\boldsymbol{\epsilon} - \mathbf{z}^{\text{tar}})
  \bigr\|_1 \bigr].
\end{equation}
At inference, the source volume is encoded to $\mathbf{z}^{\text{src}}$, the target latent is initialized as $\mathbf{z}_1 \sim
\mathcal{N}(\mathbf{0},\mathbf{I})$, and the learned ODE $\mathrm{d}\mathbf{z}/\mathrm{d}t = v_\theta(\mathbf{z}_t, t,
\mathbf{z}^{\text{src}}, m)$ is integrated from $t=1$ to $t=0$, yielding $\hat{\mathbf{z}}^{\text{tar}}$.
Because the model is conditioned on an explicit target modality token, it supports \emph{compositional} inference: given a path $A \to B \to C$ with no direct $A \to C$ supervision, we first synthesize $\hat{\mathbf{z}}^{B}$ from $\mathbf{z}^{A}$ 
conditioned on token $B$, then synthesize $\hat{\mathbf{z}}^{C}$ from $\hat{\mathbf{z}}^{B}$ conditioned on token $C$, decoding only the final output. 

\paragraph{Native-resolution training via bucketed sampling.}
Whole-volume training requires handling heterogeneous spatial dimensions without resampling all scans to a single canonical grid. 
To address this, we use a bucketed sampling strategy. 
Each volume is cropped or zero-padded so that its spatial dimensions are multiples of 128. 
Volumes with the same resulting spatial dimensions are assigned to the same bucket, so that each bucket contains only samples with identical tensor shape.
During training, each mini-batch is drawn from a single bucket.
This guarantees shape consistency within the batch and avoids any additional on-the-fly padding or interpolation. 
Since volumes are not resampled, the native voxel spacing of each acquisition is preserved; the bucketed strategy only standardizes tensor extents through cropping and zero-padding. 
As a result, the model can process whole volumes at their native physical resolution while still supporting
heterogeneous anatomical regions, modalities, and source datasets across training batches.

\section{Experiments}
\label{sec:experimental}
\noindent\textbf{Datasets.}
We evaluate on three public paired datasets spanning inter- and intra-modality translation across heterogeneous anatomical regions and acquisition conditions.
\textbf{SynthRad2023}~\cite{thummerer2023} provides brain and pelvis T1w--CT and CBCT--CT pairs, with 360 pairs per task.
\textbf{SynthRad2025}~\cite{thummerer2025} extends the same tasks to head-and-neck, thorax, and abdomen, with 578 T1w--CT and 579 CBCT--CT pairs.
\textbf{BraTS2023}~\cite{baid2021} provides 1{,}251 co-registered brain MRI cases with T1w, T1c, T2w, and T2f sequences, enabling all pairwise MRI translations.
All datasets are split 75/25 into training and test sets.
The shared modality vocabulary is $\mathcal{M}=\{\text{CT, T1w, T1c, T2w, T2f}\}$, with SynthRad T1 and BraTS native T1 assigned to the same T1w token.

\noindent\textbf{Preprocessing.}
All volumes are processed as whole 3D scans.
Before encoding, each volume is cropped to the patient foreground, thereby removing external background, and subsequently zero-padded so that its spatial dimensions are multiples of 128, up to a maximum size of $512 \times 512 \times 256$ voxels.
CT and CBCT volumes are intensity-clipped to $[-1024, 3000]$ HU and linearly mapped to $[0, 1]$.  
MRI volumes are normalised by the 99.5th percentile of each volume.

\noindent\textbf{Experimental configuration.}
We compare task-specific FM-ST Whole-Vol.\ models, trained independently for each translation task, with a unified FM-MT Whole-Vol.\ model trained jointly on SynthRad and BraTS.
For external comparison, we include SRGAN~\cite{ha2025multi}, the strongest baseline reported in the benchmark of Romoli et al.~\cite{romoli2026} and patch-based flow matching (FM-ST Patch), which processes $96^3$ patches and reconstructs full volumes
by sliding-window inference with 0.625 overlap and Gaussian blending ($\sigma$ scale $=0.125$), thus isolating the contribution of whole-volume latent processing.
Results are taken from~\cite{romoli2026} when available; otherwise, patch-based models are retrained under the same protocol.

\noindent\textbf{Evaluation protocol.}
Translation quality is assessed via Structural Similarity Index (SSIM,~$\uparrow$) and Peak Signal-to-Noise Ratio (PSNR,~$\uparrow$), computed over the foreground region defined by the patient-outline mask. 
Statistical significance is assessed via Wilcoxon signed-rank tests with Holm correction.

\noindent\textbf{Implementation details.}
The generative U-Net comprises four resolution levels with channel widths 64, 128, 256, and 512, two residual blocks per level, and self-attention at the two deepest scales. 
Training runs for 200 epochs with the Adam optimizer (lr$=10^{-4}$, polynomial decay, power 2), batch size 4, and mixed precision. 
The rectified flow scheduler uses 1{,}000 training timesteps and 30 inference steps. 
As an indicative upper bound on achievable fidelity, encoding and decoding each volume through the frozen MAISI autoencoder yields PSNR 39.93\,dB and SSIM 0.972.
All experiments were conducted on a single NVIDIA A40 GPU.

\section{Results}
\begin{table*}[t]
\centering
\caption{PSNR (dB) and SSIM results across datasets, tasks, and 
models (mean $\pm$ std). Bold: best; underlined: second best. $^*$ indicates a significant improvement of FM-MT over FM-ST Whole-Vol.\ (Wilcoxon signed-rank test with Holm correction, $p < 0.05$); all other FM-MT results are not significantly worse than FM-ST Whole-Vol.\ $^\dagger$ indicates model retrained under the protocol of~\cite{romoli2026}.}
\label{tab:psnr_ssim_results}
\resizebox{\textwidth}{!}{
\begin{tabular}{ll l l c c c c}
\toprule
\textbf{Metric} & \textbf{Dataset} & \textbf{Task} & \textbf{region}
& \textbf{SRGAN}
& \textbf{FM-ST Patch}
& \textbf{FM-ST Whole-Vol.}
& \textbf{FM-MT Whole-Vol.} \\
\midrule

\multirow{16}{*}{\rotatebox{90}{PSNR}}
& SynthRad23 & T1w$\rightarrow$CT  & Brain     
& $27.48 \pm 1.47$ 
& $24.82 \pm 0.83$ 
& $\underline{29.35 \pm 1.05}$ 
& $\mathbf{29.56 \pm 1.23}^*$ \\

& SynthRad23 & T1w$\rightarrow$CT  & Pelvis    
& $\mathbf{29.15 \pm 1.64}$ 
& $25.94 \pm 1.37$ 
& $\underline{28.74 \pm 1.97}$ 
& $28.69 \pm 1.45$ \\

& SynthRad23 & CBCT$\rightarrow$CT & Brain     
& $28.88 \pm 2.05$ 
& $25.68 \pm 1.84$ 
& $\underline{30.82 \pm 2.28}$ 
& $\mathbf{30.85 \pm 2.36}$ \\

& SynthRad23 & CBCT$\rightarrow$CT & Pelvis    
& $29.83 \pm 2.29$ 
& $24.78 \pm 1.21$ 
& $\underline{30.59 \pm 2.17}$ 
& $\mathbf{30.80 \pm 2.27}$ \\

\cmidrule(lr){2-8}

& SynthRad25 & T1w$\rightarrow$CT  & Head-Neck 
& $26.32 \pm 3.01$ 
& $21.42 \pm 1.45$ 
& $\mathbf{28.22 \pm 1.45}$ 
& $\underline{28.09 \pm 0.75}$ \\

& SynthRad25 & T1w$\rightarrow$CT  & Thorax    
& $25.74 \pm 1.51$ 
& $21.83 \pm 1.82$ 
& $\mathbf{26.79 \pm 1.89}$ 
& $\underline{26.76 \pm 1.63}$ \\

& SynthRad25 & T1w$\rightarrow$CT  & Abdomen   
& $25.91 \pm 2.18^\dagger$ 
& $21.67 \pm 1.73^\dagger$ 
& $\underline{27.26 \pm 2.59}$ 
& $\mathbf{27.43 \pm 2.31}^*$ \\

& SynthRad25 & CBCT$\rightarrow$CT & Head-Neck 
& $29.50 \pm 1.70$ 
& $26.55 \pm 1.92$ 
& $\underline{30.91 \pm 1.21}$ 
& $\mathbf{31.22 \pm 1.35}^*$ \\

& SynthRad25 & CBCT$\rightarrow$CT & Thorax    
& $28.07 \pm 1.98^\dagger$ 
& $24.90 \pm 2.12^\dagger$ 
& $\mathbf{30.57 \pm 1.69}$ 
& $\underline{30.56 \pm 1.79}$ \\

& SynthRad25 & CBCT$\rightarrow$CT & Abdomen   
& $28.35 \pm 2.10^\dagger$ 
& $25.20 \pm 2.15^\dagger$ 
& $\underline{30.82 \pm 1.85}$ 
& $\mathbf{31.05 \pm 1.92}^*$ \\

\cmidrule(lr){2-8}

& BraTS23 & T1w$\rightarrow$T1c & Brain 
& $23.84 \pm 2.71^\dagger$ 
& $22.46 \pm 2.38^\dagger$ 
& $\underline{24.96 \pm 2.33}$ 
& $\mathbf{26.51 \pm 3.42}^*$ \\

& BraTS23 & T1c$\rightarrow$T1w & Brain 
& $28.74 \pm 4.36^\dagger$ 
& $26.91 \pm 3.89^\dagger$ 
& $\mathbf{30.02 \pm 5.23}$ 
& $\underline{29.86 \pm 4.89}$ \\

& BraTS23 & T1w$\rightarrow$T2w & Brain 
& $25.96 \pm 3.01^\dagger$ 
& $24.31 \pm 2.76^\dagger$ 
& $\mathbf{27.45 \pm 3.22}$ 
& $\underline{27.31 \pm 3.25}$ \\

& BraTS23 & T2w$\rightarrow$T1w & Brain 
& $27.42 \pm 4.12^\dagger$ 
& $25.85 \pm 3.67^\dagger$ 
& $\mathbf{29.07 \pm 4.87}$ 
& $\underline{28.79 \pm 4.83}$ \\

& BraTS23 & T2w$\rightarrow$T2f & Brain 
& $25.28 \pm 2.90$ 
& $23.55 \pm 2.56$ 
& $\underline{28.92 \pm 3.17}$ 
& $\mathbf{29.12 \pm 2.94}$ \\

& BraTS23 & T2f$\rightarrow$T2w & Brain 
& $25.41 \pm 3.08^\dagger$ 
& $23.97 \pm 2.74^\dagger$ 
& $\mathbf{28.29 \pm 3.34}$ 
& $\underline{27.96 \pm 3.41}$ \\

\midrule

\multirow{16}{*}{\rotatebox{90}{SSIM}}

& SynthRad23 & T1w$\rightarrow$CT  & Brain     
& $0.84 \pm 0.04$ & $0.77 \pm 0.04$ 
& $\underline{0.86 \pm 0.02}$ & $\mathbf{0.89 \pm 0.02}^*$ \\

& SynthRad23 & T1w$\rightarrow$CT  & Pelvis    
& $\mathbf{0.88 \pm 0.02}$ & $0.73 \pm 0.05$ 
& $\underline{0.86 \pm 0.04}$ & $\underline{0.86 \pm 0.03}$ \\

& SynthRad23 & CBCT$\rightarrow$CT & Brain     
& $0.89 \pm 0.04$ & $0.78 \pm 0.09$ 
& $\mathbf{0.91 \pm 0.04}$ & $\mathbf{0.91 \pm 0.04}$ \\

& SynthRad23 & CBCT$\rightarrow$CT & Pelvis    
& $0.88 \pm 0.03$ & $0.76 \pm 0.05$ 
& $\mathbf{0.90 \pm 0.04}$ & $\mathbf{0.90 \pm 0.04}$ \\

\cmidrule(lr){2-8}

& SynthRad25 & T1w$\rightarrow$CT  & Head-Neck 
& $0.76 \pm 0.08$ & $0.53 \pm 0.05$ 
& $\mathbf{0.86 \pm 0.03}$ & $\underline{0.85 \pm 0.02}$ \\

& SynthRad25 & T1w$\rightarrow$CT  & Thorax    
& $0.73 \pm 0.05$ & $0.56 \pm 0.06$ 
& $\underline{0.77 \pm 0.08}$ & $\mathbf{0.78 \pm 0.07}$ \\

& SynthRad25 & T1w$\rightarrow$CT  & Abdomen   
& $0.74 \pm 0.07^\dagger$ & $0.55 \pm 0.06^\dagger$ 
& $\underline{0.78 \pm 0.10}$ & $\mathbf{0.80 \pm 0.09}^*$ \\

& SynthRad25 & CBCT$\rightarrow$CT & Head-Neck 
& $0.87 \pm 0.04$ & $0.71 \pm 0.09$ 
& $\underline{0.91 \pm 0.02}$ & $\mathbf{0.92 \pm 0.02}$ \\

& SynthRad25 & CBCT$\rightarrow$CT & Thorax    
& $0.80 \pm 0.05^\dagger$ & $0.65 \pm 0.11^\dagger$ 
& $\mathbf{0.88 \pm 0.03}$ & $\mathbf{0.88 \pm 0.03}$ \\

& SynthRad25 & CBCT$\rightarrow$CT & Abdomen   
& $0.81 \pm 0.06^\dagger$ & $0.66 \pm 0.10^\dagger$ 
& $\underline{0.87 \pm 0.04}$ & $\mathbf{0.89 \pm 0.03}^*$ \\

\cmidrule(lr){2-8}

& BraTS23 & T1w$\rightarrow$T1c & Brain 
& $0.86 \pm 0.04^\dagger$ & $0.78 \pm 0.05^\dagger$ 
& $\underline{0.91 \pm 0.01}$ & $\mathbf{0.92 \pm 0.01}$ \\

& BraTS23 & T1c$\rightarrow$T1w & Brain 
& $0.90 \pm 0.03^\dagger$ & $0.81 \pm 0.05^\dagger$ 
& $\mathbf{0.93 \pm 0.02}$ & $\mathbf{0.93 \pm 0.01}$ \\

& BraTS23 & T1w$\rightarrow$T2w & Brain 
& $0.88 \pm 0.04^\dagger$ & $0.78 \pm 0.05^\dagger$ 
& $\mathbf{0.92 \pm 0.01}$ & $\mathbf{0.92 \pm 0.01}$ \\

& BraTS23 & T2w$\rightarrow$T1w & Brain 
& $0.90 \pm 0.03^\dagger$ & $0.82 \pm 0.05^\dagger$ 
& $\mathbf{0.93 \pm 0.02}$ & $\mathbf{0.93 \pm 0.02}$ \\

& BraTS23 & T2w$\rightarrow$T2f & Brain 
& $0.83 \pm 0.05$ & $0.59 \pm 0.06$ 
& $\underline{0.88 \pm 0.01}$ & $\mathbf{0.89 \pm 0.01}$ \\

& BraTS23 & T2f$\rightarrow$T2w & Brain 
& $0.87 \pm 0.04^\dagger$ & $0.74 \pm 0.05^\dagger$ 
& $\mathbf{0.90 \pm 0.01}$ & $\mathbf{0.90 \pm 0.01}$ \\

\bottomrule
\end{tabular}
}
\end{table*}

\paragraph{Whole-volume vs.\ patch-based processing.}
Table~\ref{tab:psnr_ssim_results} compares all evaluated models across tasks, datasets, and anatomical regions. 
FM-ST Patch yields the lowest performance across all evaluated tasks, confirming that patch-based flow matching without a strong volumetric prior fails to exploit the generative architecture's full capacity. 
SRGAN improves substantially over FM-ST Patch, yet is outperformed by both whole-volume models on all tasks except Pelvis T1w$\to$CT, where it retains a narrow advantage in PSNR 
and SSIM. 
Across all tasks and datasets, FM-ST Whole-Vol.\ outperforms FM-ST Patch, with gains in SSIM ranging 
from $+0.12$ to $+0.37$ and in PSNR from $+2.8$ to $+8.4$\,dB. 
These results empirically validate the central claim of this work: when a strong volumetric prior makes whole-volume processing tractable, operating on full volumes is preferable to patch-based 
inference at parity of generative architecture.

\begin{figure}[t]
  \centering
  \includegraphics[width=\textwidth]{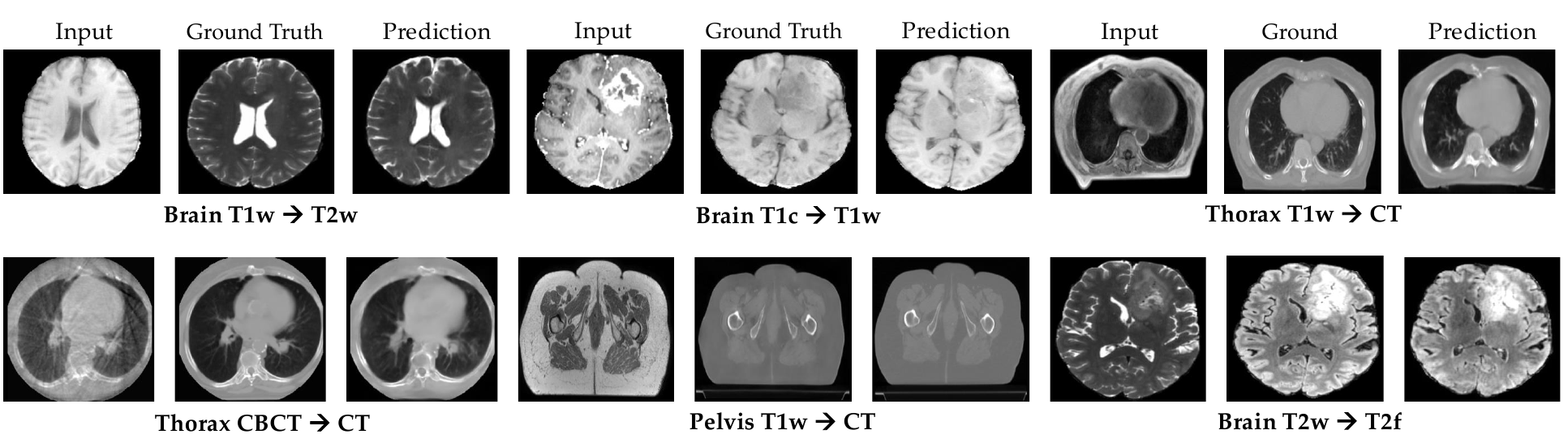}
  \caption{Representative qualitative results across inter-modality and intra-modality
translation tasks. For each example, input, ground truth, and prediction are shown.}
  \label{fig:qualitative}
\end{figure}

\paragraph{Single-task vs.\ multitask training.}
We compare the task-specific FM-ST Whole-Vol.\ models with the unified FM-MT Whole-Vol.\ model across all evaluated datasets, translation tasks, and anatomical regions. 
FM-MT is never significantly worse than the corresponding single-task baseline on any evaluated metric. 
Significant improvements over FM-ST Whole-Vol.\ are marked with $^*$ in Table~\ref{tab:psnr_ssim_results} and indicate cases in which joint training provides a measurable benefit. 
These results show that a single multitask model can replace $N$ independently trained task-specific networks without a detectable loss in task-level accuracy.
Moreover, the significant gains observed in several settings suggest that joint training can act as a beneficial regularizer by sharing information across anatomically and modality-overlapping translation tasks. 
The benefits of unification extend beyond pixel-level accuracy, as shown by the generalization and compositional experiments reported next. 
Representative results are shown in Fig.~\ref{fig:qualitative}.

\paragraph{Leave-One-Region-Out (LORO).}
To assess zero-shot anatomical generalization, we train five FM-MT models, each excluding one anatomical region, and evaluate on the held-out region. 
Table~\ref{tab:lodo} reports SSIM averaged over T1w$\to$CT and CBCT$\to$CT for each excluded region. 
The model reaches an average SSIM of 0.72 against 0.87 for the full multitask model.
The limited degradation across all regions confirms that joint training over the full dataset collection induces a shared latent geometry rich enough to support partial anatomical generalization.

\begin{table}[t]
\caption{Leave-One-Region-Out (LORO) evaluation. SSIM values
are averaged over T1w$\to$CT and CBCT$\to$CT tasks. ``Full''
denotes the multitask model trained on all regions; ``LORO''
denotes zero-shot performance on the excluded region.}
\label{tab:lodo}
\centering
\small
\setlength{\tabcolsep}{5pt}
\begin{tabular}{lccc}
\toprule
region & Full & LORO & $\Delta$ \\
\midrule
Brain     & 0.90 & 0.74 & $-$0.16 \\
Pelvis    & 0.88 & 0.71 & $-$0.17 \\
Abdomen   & 0.84 & 0.72 & $-$0.12 \\
Head-Neck & 0.89 & 0.72 & $-$0.17 \\
Thorax    & 0.83 & 0.70 & $-$0.13 \\
\midrule
Average   & 0.87 & 0.72 & $-$0.15 \\
\bottomrule
\end{tabular}
\end{table}

\subsection{Compositional Translation via Two-Stage Inference}
We evaluate the compositional capabilities of the Multitask model through two-stage inference. 
We consider two scenarios of increasing complexity.

\paragraph{Intra-dataset chaining.}
As a first validation, we evaluate T1w$\to$T2w$\to$T2f on BraTS, comparing FM-MT Whole-Vol.\ with a chain of two FM-ST Whole-Vol.\ models. 
This supervised setting serves to characterize error propagation across sequential translations. 
At stage~1, both approaches achieve comparable performance. 
At stage~2, however, FM-MT improves over the single-task chain (SSIM~0.95 vs.\ 0.92, $p < 0.05$, Table~\ref{tab:twostage_brats}).
This suggests that the intermediate T2w generated by the multitask model remains better aligned with the subsequent T2w$\to$T2f mapping. 
In contrast, the single-task chain provides the second model with a synthetic T2w produced by an independently trained model, creating a mismatch with the real T2w inputs used during its training.

\begin{figure}[t]
  \centering
  \includegraphics[width=0.7\textwidth]{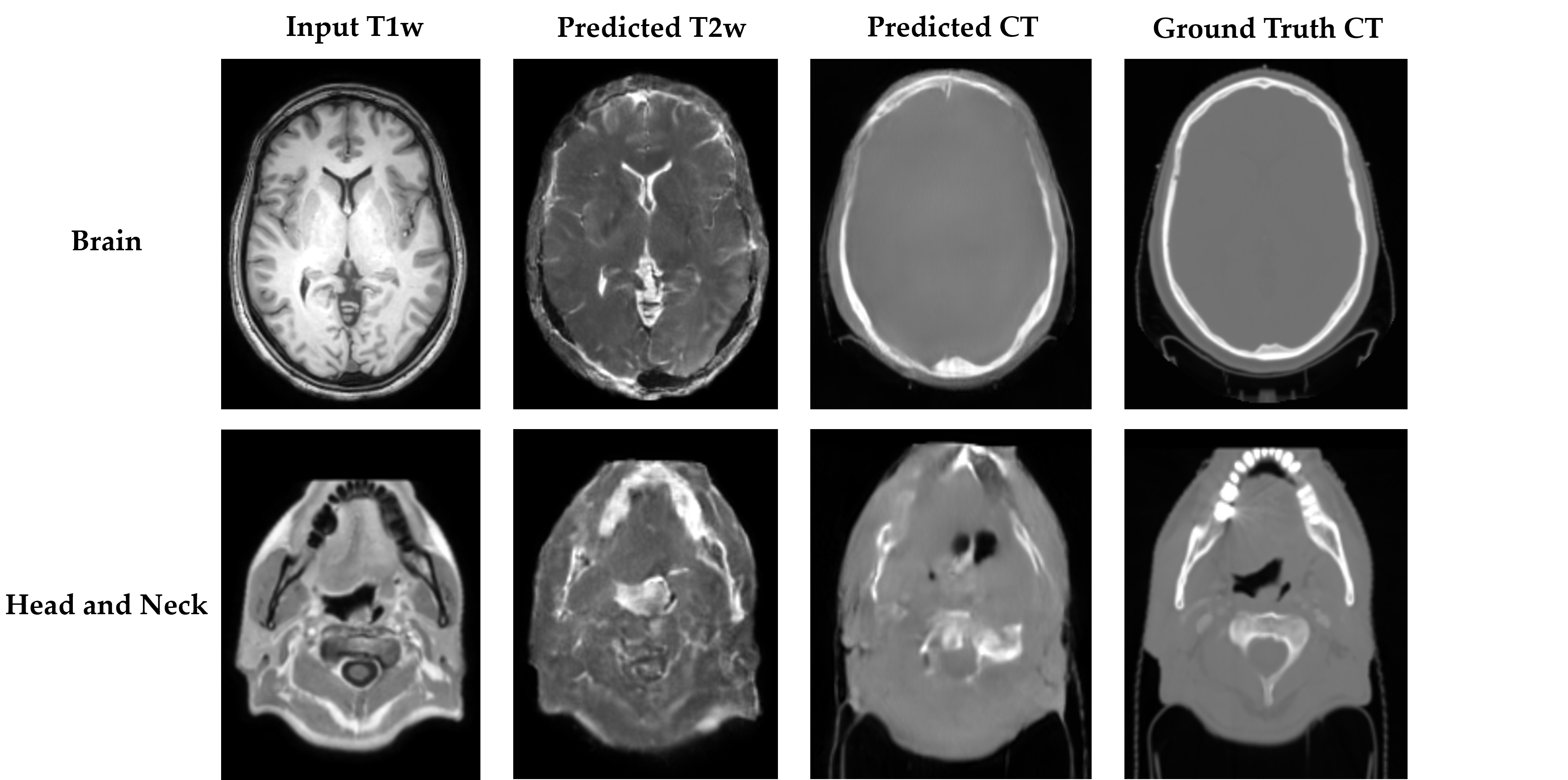}
  \caption{Cross-dataset compositional translation (T1w$\to$T2w$\to$CT). Brain (top) and head-and-neck
  (bottom).}
  \label{fig:twostage}
\end{figure}

\paragraph{Cross-dataset chaining.}
We then evaluate a more challenging cross-dataset path, T1w$\to$T2w$\to$CT, where the first step exploits BraTS knowledge and the second step targets SynthRad CT synthesis. 
This path is never directly supervised, since no T2w$\to$CT pairs are available.
We report results on brain and head-and-neck, the regions with the closest anatomical overlap with BraTS.
As shown in Table~\ref{tab:twostage_synth}, FM-MT achieves SSIM 0.84 and 0.80 on brain and head-and-neck, respectively, while the chained FM-ST baseline drops to 0.15 and 0.31.
This collapse reflects the distribution mismatch between the synthetic T2w generated by the BraTS-only model and the inputs expected by
the SynthRad-only model, which was never trained on T2w representations.
Representative examples are shown in Fig.~\ref{fig:twostage}.

\begin{table}[!t]
  \centering
  \begin{minipage}{0.52\textwidth}
    \centering
    \caption{Two-stage intra-dataset chaining on BraTS 
    (T1w$\to$T2w$\to$T2f). Statistically significant improvement 
    at stage~2 ($p < 0.05$, Wilcoxon with Holm correction).}
    \label{tab:twostage_brats}
    \resizebox{\linewidth}{!}{
    \begin{tabular}{lcccc}
    \toprule
    & \multicolumn{2}{c}{Stage 1: T1w$\to$T2w}
    & \multicolumn{2}{c}{Stage 2: T2w$_{\text{pred}}\to$T2f} \\
    \cmidrule(lr){2-3}\cmidrule(lr){4-5}
    Model & SSIM & PSNR & SSIM & PSNR \\
    \midrule
    FM-ST (chained)
      & \textbf{0.96} & \textbf{28.45}
      & 0.92 & 28.87 \\
    FM-MT (Multitask)
      & \textbf{0.96} & 28.32
      & \textbf{0.95} & \textbf{30.80} \\
    \bottomrule
    \end{tabular}}
  \end{minipage}
  \hfill
  \begin{minipage}{0.45\textwidth}
    \centering
    \caption{Two-stage cross-dataset chaining 
    (T1w$\to$T2w$_{\text{BraTS}}\to$CT on SynthRad).}
    \label{tab:twostage_synth}
    \resizebox{\linewidth}{!}{
    \begin{tabular}{lcccc}
    \toprule
    & \multicolumn{2}{c}{Brain}
    & \multicolumn{2}{c}{Head-Neck} \\
    \cmidrule(lr){2-3}\cmidrule(lr){4-5}
    Model & SSIM & PSNR & SSIM & PSNR \\
    \midrule
    FM-ST (chained)
      & 0.15 & 12.88
      & 0.31 & 13.91 \\
    FM-MT (Multitask)
      & \textbf{0.84} & \textbf{25.51}
      & \textbf{0.80} & \textbf{25.82} \\
    \bottomrule
    \end{tabular}}
  \end{minipage}
\end{table}

\section{Discussion and Conclusion}
This work shows that whole-volume latent translation improves over patch-based inference, and multitask training can replace task-specific models without significant degradation. 
Beyond accuracy, the unified model enables zero-shot anatomical generalization and compositional translation paths that are not supported by independently trained models.
Limitations remain. 
Evaluation is based on pixel-level metrics and does not yet establish clinical utility. 
Future work should assess downstream relevance and expert reader studies. 
The method also inherits the modality and anatomical coverage of the pretrained VAE prior, and zero-shot region generalization remains below fully supervised performance.
Future work will therefore focus on broader priors, few-shot adaptation, and uncertainty estimation for out-of-distribution regions.

\bibliographystyle{splncs04}
\bibliography{references}

\end{document}